# Using LLVM-based JIT Compilation in Genetic Programming

Michal Gregor*, Juraj Spalek†
Department of Control and Information Systems
Faculty of Electrical Engineering
University of Žilina, Žilina, Slovak Republic
Email: *michal.gregor@fel.uniza.sk, †juraj.spalek@fel.uniza.sk

*Abstract*—The paper describes an approach to implementing genetic programming, which uses the LLVM library to just-in-time compile/interpret the evolved abstract syntax trees. The solution is described in some detail, including a parser (based on FlexC++ and BisonC++) that can construct the trees from a simple toy language with C-like syntax. The approach is compared with a previous implementation (based on direct execution of trees using polymorphic functors) in terms of execution speed.

## I. INTRODUCTION

Genetic programming is a well-known approach to automatic evolution of computer programs. It has often been applied to tasks that involve symbolic regression, automatic creation of algorithmic solutions, rule learning and other similar tasks. Genetic programming has proved its usefulness on many problems. There is a number of tools available in many languages that support genetic programming.

One of the issues in genetic programming is the efficiency of execution. In many tasks, such as symbolic regression, the evolved program has to be executed many times. In such cases, execution of the solution will easily dominate the total execution time. In such applications then, if we are to improve the computational efficiency of genetic programming, we need to focus on making the execution of the evolved programs as fast as possible.

The present paper explores one possible way to do that: we show how to use the LLVM library (the collection of language-agnostic tools used in the Clang compiler) to JIT compile/interpret the evolved program. We also compare this solution with our previous approach – based on direct execution of the tree using polymorphic functors – in terms of execution speed.

The rest of the paper is organized as follows. Section II. will consider some basics concerning genetic programming so as to provide the necessary context for the other sections. Section III. will briefly present the main features of our original GP solution, based on direct execution with polymorphic functors – so as to provide some context for the approach proposed in the present paper.

Section IV. will discuss how just-in-time compilation/interpretation of evolved programs (expressed in the form of abstract syntax trees) can be accomplished using the LLVM library. Section V. shows how abstract syntax trees can be created by parsing sources in a simple toy language with C-like syntax. This is especially useful when we need to bootstrap evolution using hand-crafted solutions.

In section VI. we will give a description of the evaluation procedure used to measure the execution times. Finally, the actual empirical results will be given in section VII. for several variants of the proposed solution. The results will give valuable indications as to how many times a solution has to be run before the performance-related benefits of JIT compilation and optimization will outweigh the additional time consumed by those processes themselves.

## II. CONCERNING GENETIC PROGRAMMING

Genetic programming (GP) is a well-known technique for automatic evolution of computer programs. The foundations have been laid by John Koza [1]. The main idea behind genetic programming is to use abstract syntax trees (ASTs) to represent programs. This makes implementation of genetic operators relatively straight-forward. Generally speaking, crossover relies on swapping subtrees between parents and mutation replaces randomly selected subtrees.

A further advantage of using ASTs as a representation is that there is a large amount of existing methods and tools that work nicely with them. This is because ASTs are the conventional representation used in parsers and compilers.

### A. Existing Approaches to Compiled GP

In this subsection we will present a brief overview of some existing approaches to using compilation in genetic programming. Most approaches that have attempted the use of some form of compilation in genetic programming have switched to linear representations as opposed to using the ASTs.

For an instance, in [2], [3] the authors have used virtual stack machines to execute the evolved programs. The programs themselves are composed of instructions for such stack machines – e.g. push instructions, no operation instructions, basic arithmetic and boolean operations. This is a fairly low-level representation, which makes it more difficult to analyse. It also impairs interpretability to an extent. However, the implementation of genetic operators is still straight-forward

– the standard operators for manipulating strings can be used. A considerable limitation in [2] is the lack of support for branching, although this is remedied in [3].

A further interesting approach – called the compiling genetic programming system – was presented by Nordin in [4] – this one directly manipulates machine code instructions. (Despite the name there is no separate compilation procedure.)

A further approach proposed in [5] uses ASTs as the internal representation, but translates them into linear instructions for a virtual stack machine. This combines the advantages of using ASTs as a representation and also allows for faster execution using a stack machine. However, it involves a separate compilation procedure because it has to translate between the two representations. Paper [5] also discusses the problem of executing many solutions in parallel.

There is also the approach of Fukunaga et al. [6], which combines some of the advantages from [4] and [5] by compiling ASTs as [5], but not into a representation for a virtual stack machine, but directly using machine code as [4] does.

The two last-mentioned approaches are very similar to what we seek to accomplish. The principal difference is that our method uses the LLVM library, which allows us to easily provide support for further advanced features such as automatic code optimization. We are also able to compile the solutions into native code for a large number of platforms. In addition, we are able to support both – just-in-time compilation and interpretation – using a single codebase.

The situation has changed significantly in the recent years – there have been several major technical developments, not the least of which was the introduction of the LLVM library and other accompanying tools. When Poli et al. mention the possibility of compiling GP programs in their Field Guide to Genetic Programming [7], they conclude that while it is certainly possible to compile the evolved programs, writing a compiler involves a lot of work and so it is more common to use some kinds of interpreters to execute programs. With the availability of LLVM, the amount of work required in order to apply just-in-time compilation to the evolved ASTs, has been reduced substantially.

### III. DIRECT POLYMORPHIC EXECUTION

In a later section we will be comparing our original GP implementation – based on direct execution of the AST with polymorphic functors – with the proposed solution based on just-in-time compilation/interpretation in terms of execution times. It is therefore requisite to include some basic information concerning our original implementation so that we have a good idea of what we are comparing.

We will focus mostly on the representation of the AST itself – this is based on class `Tree`. Function signatures are specified using `TypeList` and `FunctorSignature` objects. Type identifiers are derived from our own portable system based on templates and macros at compile time.

A `Tree` object contains a pointer to the root `Node` of the AST. `Node` objects are themselves non-virtual – but they contain NodeFunctor objects, which are polymorphic and implement the actual behaviour of the corresponding nodes. There is a considerable downside to this approach – the ASTs are opaque and they cannot easily be processed (e.g. from some optimization passes). On the other hand, this disadvantage could be remedied relatively easily.

The good thing about this approach is that the ASTs can be run as they are – there is no need to translate them into any intermediate representation first. Issues concerning memory allocation for local variables and such, are handled using context blocks – the details can be found in [8]. Some further information – concerning the application of contexts and context blocks to the evolution of constants and modules is provided in [9], [10]. We have already discussed conceptually how these applications could be transitioned to the LLVM-based system in [11] – however there was no implementation of such a solution at the time.

In any case – although a further translation step is necessary prior to execution if we use LLVM, the execution itself is faster. It is also quite cumbersome to be forced to implement all runtime logic under the direct approach – since it necessitates the use of context blocks. It would, for an instance, be very difficult (and computationally expensive) to implement proper scoping for variables in this way. LLVM has built-in support for many such concepts.

### IV. JUST-IN-TIME COMPILATION OF ASTS

The above-mentioned disadvantages of the direct approach have motivated us to propose an alternative solution, which uses the LLVM library [12] to just-in-time (JIT) compile/interpret the ASTs. This approach requires that each AST be translated into intermediate code (and optionally JIT compiled) before it is executed. This comes at a computational cost. However, the actual execution of the AST, once translated, is faster. One of our objectives in this paper is to provide some indication as to how many times a typical AST would have to be run before these performance improvements would outweigh the translation costs.

A further advantage to using LLVM in place of our previous approach is its robustness and easy out-of-box support for many advanced features. LLVM is used in production code and its use as a part of the Clang compiler and other high-profile projects gives strong quality guarantees. Using LLVM allows us to dispense with the use of ProcessContext objects altogether. Naturally, these advantages hold even in cases where execution times are not improved by using LLVM. They form a compelling (though perhaps not a decisive) reason for selecting LLVM in such settings as well.

#### A. The Translation Procedure

The entire translation procedure is as follows:
1) The original AST is translated into LLVM's IR – intermediate representation, which is somewhat similar to an assembly language.
2) The intermediate code is (optionally) processed using some optimization passes (constant folding, instruction combination, etc.).

3) The IR is then either:
   - just-in-time compiled,
   - interpreted on the fly.

In order to perform the initial translation from the AST into IR, we use LLVM's C++ interface. Once the translation process is completed, the subsequent steps are all managed by LLVM – we merely need to direct the process by selecting some appropriate optimization passes, targets, and so on.

As we can see, if we use LLVM interpretation or just-in-time compilation, the translation will in any case involve at least one additional pass over the AST – instead of executing it directly, we first have to translate it into IR and then execute that. If the resulting intermediate code is to be optimized, even more passes may be necessary.

This indicates that at least with very few repetitions, the proposed LLVM-based solution may have a hard time competing with our previous direct approach. It will be shown hereinafter that although the LLVM-based solution is significantly faster if the AST has to be executed multiple times, there is a performance penalty for ASTs that are only run once. We have some ideas as to how the overhead could be reduced, but it is not clear that the LLVM-based version will even then be able to beat the direct approach in single runs.

### B. The Structure of the AST

In Fig. 1 we attach a UML class diagram, which illustrates the class hierarchy used to represent the ASTs. As we can see in the figure, the trees are composed of nodes, which can be of various kinds. Similarly to standard programming languages, there are two basic categories – expressions and statements.

Every one of the classes has a virtual `codegen()` function, which takes in LLVM's `CodeGenContext&` as a parameter. It is responsible for generating the IR corresponding to the node. For an instance, the `codegen` for `NInteger` can be implemented using a simple single-line LLVM call as follows:

```
Value* NInteger::codeGen(CodeGenContext& context) {
        return ConstantInt::get(Type::getInt64Ty(
                getGlobalContext()), value, true);
}
```

This creates an integer constant with the specified value, given the context (so that LLVM knows where to output the IR and so on).

### C. Calling Native Code

Calls to native C functions from the LLVM IR are also possible – it is merely necessary to generate the appropriate *extern* declaration into the IR first and to allow dynamic linking. With C++ functions and methods, the matter is a bit more difficult – we would have to wrap every one of them within an `extern "C"` scope in order to make them work using the extern declaration.

There is, however, a way around that – there are ways to specify the address of a symbol manually. (In LLVM 3.6 this can be accomplished by deriving from the `SectionMemoryManager` – this works for the JIT engine; or `addGlobalMapping()` – this works for the

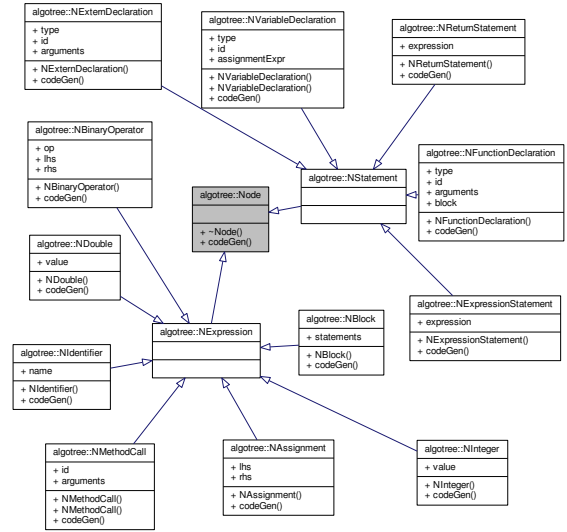

Fig. 1: A UML class diagram of AST node types.

`Interpreter` engine) Using such options, we can create a look-up table, which will map function names to function addresses. This way we can call all standard C++ functions, including even templates. That, in its turn, makes it possible to call member functions by automatically wrapping them in variadic template wrappers (even though the cost of each call will, in consequence, increase slightly).

Although this can be done, we have not conducted any extensive testing of the feature and the performance of native calls is not evaluated in this paper. We hope to conduct more detailed testing of this aspect as part of our future work.

## V. CREATING ASTs USING A PARSER

In addition to LLVM-based JIT compilation/interpretation we have also implemented a simple parser, which can create ASTs from human-readable source code. This was necessary for evaluating the LLVM-based JIT compilation/interpretation – we needed to manually construct ASTs so as to measure their execution times. Being able to easily manually construct ASTs is also useful in practice – we may want to initialize the population using several hand-crafted solutions.

When parsing a source file, the typical procedure is to first apply a lexer. The task of the lexer is to split the original source file into tokens (e.g. numbers, strings, keywords, ...). Once we have the tokens, they are passed to the actual parser, which constructs an abstract syntax tree from them.

Our particular implementation of the lexer is based on FlexC++ [13]. FlexC++ is a tool able to automatically generate a lexer, given an input file that mostly consists of regular expressions specifying what the tokens should look like. The source code of the resulting lexer is in C++.

In a similar way, the parser is automatically generated using BisonC++ [14]. The input of BisonC++ mostly consists of a context-free grammar of the language to be parsed. The particular language used in our experiments is a simple toy language with C-like syntax. An example program is given in Listing 1.

At the moment, we do not have a way to automatically visualize the resulting abstract syntax tree. We intend to eventually provide support for automatic visualization using Graphviz [15]. However, this part of our system has not been fully implemented yet.

```
int compute(int A, int B) {
        return B + 2*A
}

int A1 = (5 + 7) * 14
int A2 = compute(A1, 11)

return A2 + compute(11, 12 + 2)
```

Listing 1: Example of the toy language source code.

We will not dwell on further details concerning the implementation in the present paper, as that would necessitate the inclusion of the regular expressions and the context-free grammar. These would take up an excessive amount of space. In any case, such implementation details seem to be rather out of the scope of the paper.

## VI. THE EVALUATION PROCEDURE

The evaluation procedure is laid out according to pseudocode in Fig. 2. We have used the short sample code from Fig. 1 as a reference example. In order to model the effects of larger code samples, we execute the same piece of code multiple times (without reparsing and recompilation). This is accomplished using the inner loop. The number of repetitions of this inner loop will be referred to as the number of *inner repetitions* hereinafter.

### A. Inner and Outer Repetitions

We also need to measure the effects of repeating the entire process – with recompilation before each trial. This is accomplished using the outer loop in Fig. 2. This number of repetitions will be referred to as the number of *outer repetitions*.

When applying GP to symbolic regression, the number of outer repetitions might correspond to the number of individuals in the population and the number of inner repetitions to the size of the dataset. This is because every individual's tree would need to be compiled once and then run for every item in the dataset.

In order to provide a comparison with our original direct implementation, we have assembled the same piece of code by hand in a form accessible to it. To give the reader some idea of how inconvenient this procedure is – and why there is a need for a parser – we will mention that assembling the AST for Fig. 1 by hand takes roughly 100 lines of code.

Note also that in neither case is the time consumed by creating the AST – whether by parsing or by manually assembling – included in the measured total execution time. The AST is created before we start measuring in Fig. 2.

Create the AST (by parsing a source or manually);
**for** $r = \{1, 2, ..., \text{averaging\_repetitions}\}$ **do**
    Start measuring the execution time;
    **for** $i = \{1, 2, ..., \text{outer\_repetitions}\}$ **do**
        Translate the AST;
        **for** $j = \{1, 2, ..., \text{inner\_repetitions}\}$ **do**
            Execute the program;
        **end**
    **end**
    Stop measuring the execution time;
**end**
Average the results from all averaging\_repetitions runs;

**Fig. 2:** Pseudocode of the evaluation procedure.

### B. Averaging the Results

Note also that in every case the measurement is repeated averaging\_repetitions times and the results are averaged over all the runs. In our particular case we used averaging\_repetitions = 50.

### C. The Tested Configurations

We have tested and compared several configurations. Broadly speaking, we are comparing the original direct approach as described in section III against two variants of the LLVM-based approach – one using JIT compilation and the other using LLVM's interpreter. Furthermore there is the choice of including various optimization passes. In our experiments we have selected the following reference collection of passes: *(a)* the basic alias analysis pass, *(b)* the instruction combining pass, *(c)* the reassociate pass, *(d)* the GVN pass, *(e)* the CFG simplification pass. In the results we also compare the JIT and the interpreter-based solutions with or without these optimization passes.

All in all this yields the following 6 configurations:
- ALG: the original direct approach;
- INT: the interpreter without optimization;
- INT-OPT: the interpreter with optimization;
- JIT: just-in-time compilation without optimization;
- JIT-OPT: just-in-time compilation with optimization.

## VII. EMPIRICAL RESULTS

It is naturally the case that the results vary significantly between the case where there is only a single inner repetition and the case when there are several. With a single inner repetition our original direct approach performs best, as shown in Fig. 3. It is followed by the interpreter without any optimization passes, but there is a noticeable gap between the two.

Unsurprisingly, when there is a larger number of inner repetitions, the resulting execution times are in sharp contrast to these results – the benefits of LLVM can then easily outweigh the additional computation required for the translation and JIT compilation/interpretation. This is verified by the results shown in Fig. 4. The entire set of results is also included in tabular form in TABLE I so that readers can make more detailed comparisons if they are inclined to.

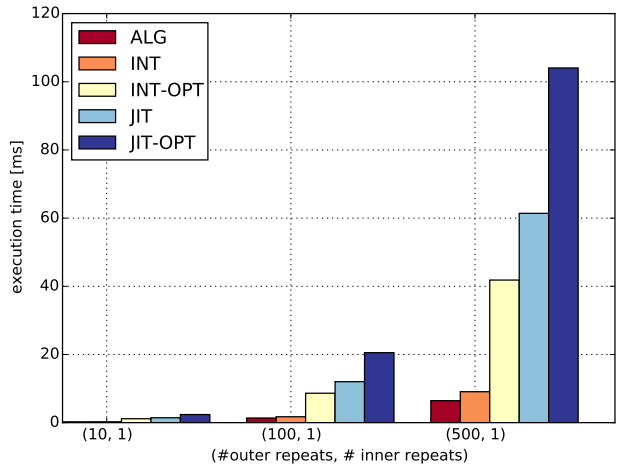

Fig. 3: Execution times with a single inner repeat.

The results indicate that the LLVM-based approach performs significantly better than direct execution (our previous approach). We can also conclude that whenever JIT compilation is applied, it should be applied in conjunction with optimization. However, at least in our current implementation and with the numbers of repetitions that we tested with, the LLVM interpreter consistently outperforms the version based on JIT compilation.

All-in-all the interpreted version with optimization seems to perform best (if there is a larger number of inner repetitions). This holds unless the number of inner repetitions approaches 10 or less, in which case the interpreter without optimization performs better.

Naturally, all of these result may also depend on the particular code samples used – especially on their length, but also on the actual quality of the code, since these may determine how much potential for optimization there is.

## VIII. CONCLUSION

Genetic programming is often applied to tasks where the evolved solution has to be rerun multiple times in order to evaluate it (e.g. symbolic regression). It is therefore essential that the execution time of the evolved AST be as short as possible. In the present paper we have proposed a GP system based on the LLVM library. LLVM allows us to just-in-time compile or interpret the AST and it also enables us to run optimization passes, which may result in further performance improvements.

We have also empirically verified that the approach performs significantly better than a simpler implementation based on direct execution of the AST using polymorphic functors. The direct implementation is, however, faster when we only need to run the AST once (or very few times). We can also conclude that, at least in our present experiments, the LLVM interpreter-based version consistently outperforms the version that uses just-in-time compilation.

All-in-all the interpreted version with optimization seems to perform best (if there is a larger number of inner repetitions). This holds unless the number of inner repetitions approaches 10 or less, in which case the interpreter without optimization performs better.

The LLVM-based approach also has further advantages. These include especially LLVM's robustness and out-of-box support for many advanced features, which allows us to dispense with the use of contexts and context blocks for mere execution. All of these benefits are present even in cases where the actual execution times are not improved by using LLVM.

In our future work we would like to identify an even more light-weight LLVM solution, which would allow us to match the performance of direct execution even in settings, where the syntax tree is executed only a few times. This would allow us to dispense with our previous GP implementation altogether. However, it is not yet clear whether this is feasible and how precisely it is to be accomplished. For one thing, mere translation to LLVM IR seems to take nearly as much time as a single direct execution.

Furthermore, in the present work we have omitted empirical evaluation of the execution time consumed using various kinds of native function calls. This also needs to be studied in some detail so that we are able to make informed decisions as to which kind of implementation would be most suitable for a particular task.

Finally, once it is completely finished, we would like to benchmark the LLVM-based implementation against other existing GP implementations – whether compilation-based or otherwise. It will likely not be feasible to produce a comparison with all the listed existing approaches to compiled GP, since their implementations are not readily available. However, there is a number of different GP implementations that are.

This contribution/publication is the result of the project implementation: **Centre of excellence for systems and services of intelligent transport**, ITMS 26220120050 supported by the Research & Development Operational Programme funded by the ERDF.

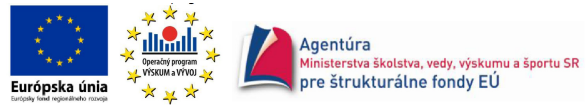

We support research activities in Slovakia / Project is cofinanced from EU funds.

## REFERENCES

[1] J. R. Koza, *Genetic Programming: On the Programming of Computers by Means of Natural Selection*. Cambridge, Massachusetts: MIT Press, 1998.
[2] T. Perkis, "Stack-based genetic programming," in *Evolutionary Computation, 1994. IEEE World Congress on Computational Intelligence., Proceedings of the First IEEE Conference on*. IEEE, 1994, pp. 148–153.
[3] K. Stoffel and L. Spector, "High-performance, parallel, stack-based genetic programming," in *Proceedings of the 1st annual conference on genetic programming*. MIT Press, 1996, pp. 224–229.

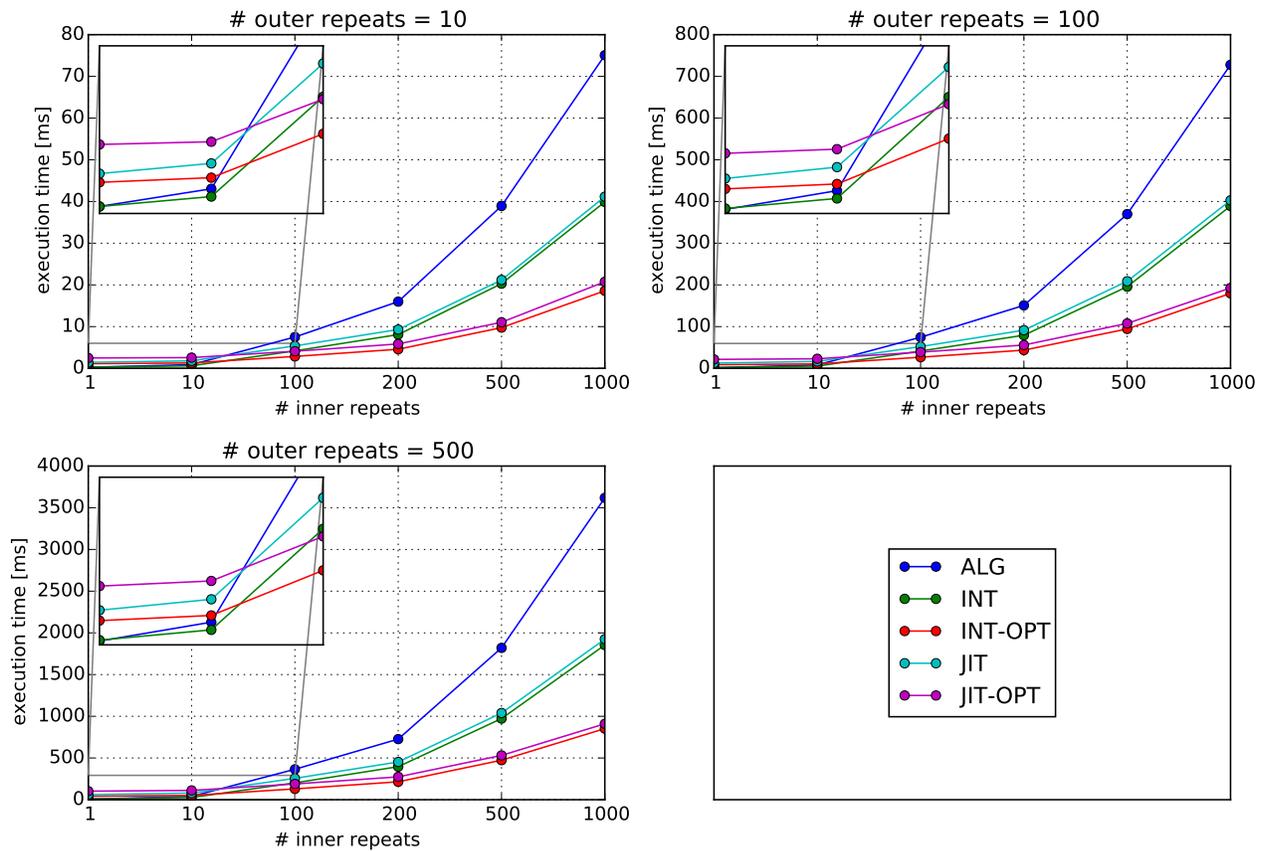

Fig. 4: Execution times in various configurations.

TABLE I: Tabular results – average executions times in milliseconds.

| outer repeats | 10 | | | | | | 100 | | | | | | 500 | | | | | |
|---|---|---|---|---|---|---|---|---|---|---|---|---|---|---|---|---|---|---|
| inner repeats | 1 | 10 | 100 | 200 | 500 | 1000 | 1 | 10 | 100 | 200 | 500 | 1000 | 1 | 10 | 100 | 200 | 500 | 1000 |
| ALG | 0.25 | 0.88 | 7.50 | 15.99 | 38.93 | 75.07 | 1.44 | 8.11 | 74.50 | 150.88 | 369.79 | 727.70 | 6.57 | 39.48 | 364.70 | 726.41 | 1820.00 | 3618.64 |
| INT | 0.25 | 0.60 | 4.18 | 8.09 | 20.30 | 39.90 | 1.79 | 5.33 | 41.42 | 79.32 | 195.80 | 389.13 | 8.25 | 26.08 | 201.97 | 395.68 | 972.29 | 1854.13 |
| INT-OPT | 1.12 | 1.28 | 2.85 | 4.57 | 9.76 | 18.58 | 8.74 | 10.48 | 26.63 | 43.57 | 94.46 | 179.69 | 42.18 | 51.19 | 129.64 | 214.73 | 473.52 | 852.01 |
| JIT | 1.43 | 1.79 | 5.37 | 9.29 | 21.18 | 41.15 | 12.44 | 16.47 | 52.08 | 91.34 | 208.73 | 402.77 | 60.26 | 79.27 | 256.00 | 452.17 | 1039.06 | 1924.31 |
| JIT-OPT | 2.47 | 2.57 | 4.09 | 5.84 | 11.04 | 20.75 | 21.38 | 22.87 | 38.86 | 55.78 | 107.70 | 192.65 | 102.27 | 111.20 | 188.75 | 273.86 | 530.88 | 909.85 |

[4] P. Nordin, "A compiling genetic programming system that directly manipulates the machine code," *Advances in genetic programming*, vol. 1, pp. 311–331, 1994.
[5] H. Juille and J. B. Pollack, "Massively parallel genetic programming," *Advances in Genetic Programming 2*, pp. 339–357.
[6] A. Fukunaga, A. Stechert, and D. Mutz, "A genome compiler for high performance genetic programming," *Genetic Programming*, pp. 86–94, 1998.
[7] R. Poli, W. B. Langdon, N. F. McPhee, and J. R. Koza, *A field guide to genetic programming*. Lulu. com, 2008. [Online]. Available: http://www.gp-field-guide.org.uk
[8] M. Gregor, J. Spalek, and J. Capák, "Use of context blocks in genetic programming for evolution of robot morphology," in *Proceedings of 9th International Conference, ELEKTRO 2012*, 2012.
[9] M. Gregor and J. Spalek, "Using context blocks to implement node-attached modules in genetic programming," in *Intelligent Engineering Systems (INES), 2013 IEEE 17th International Conference on*. IEEE, 2013, pp. 317–322.
[10] ——, "On use of node-attached modules with ancestry tracking in genetic programming," in *11th IFAC/IEEE International Conference on Programmable Devices and Embedded Systems*, 2012.
[11] ——, "Using context blocks in genetic programming with jit compilation," *ATP Journal PLUS*, no. 2, 2013.
[12] "The llvm compiler infrastructure." [Online]. Available: http://llvm.org
[13] F. B. Brokken, J.-P. van Oosten, and R. Berendsen. Flexc++ user guide. [Online]. Available: http://flexcpp.sourceforge.net/flexc++.html
[14] F. B. Brokken. Bisonc++ user guide. [Online]. Available: http://bisoncpp.sourceforge.net/bisonc++.html
[15] Graphviz – graph visualization software. [Online]. Available: http://www.graphviz.org/